\documentclass[runningheads]{llncs}
\usepackage{amsmath}
\usepackage{graphicx}
\usepackage{booktabs, multirow}
\usepackage[table]{xcolor}
\usepackage{changepage,threeparttable}
\usepackage{numprint}
\usepackage{float}
\begin{document}
\title{Introducing Inter-Relatedness between Wikipedia Articles in Explicit Semantic Analysis}
%
\titlerunning{Introducing Inter-Relatedness in ESA}
%
\author{Naveen Elango\inst{1}\orcidID{ME16B077} \and
Pawan Prasad K\inst{1}\orcidID{ME16B179}}
%
%
\institute{Indian Institute of Technology, Madras\\
\email{\{me16b077,me16b179\}@smail.iitm.ac.in}}
\maketitle              
\begin{abstract}
Explicit Semantic Analysis (ESA) is a technique used to represent a piece of text as a vector in the space of concepts, such as Articles found in Wikipedia. We propose a methodology to incorporate knowledge of Inter-relatedness between Wikipedia Articles to the vectors obtained from ESA using a technique called Retrofitting to improve the performance of subsequent tasks that use ESA to form vector embeddings. Especially we use an undirected Graph to represent this knowledge with nodes as Articles and edges as inter relations between two Articles. Here, we also emphasize how the ESA step could be seen as a predominantly bottom-up approach using a corpus to come up with vector representations and the incorporation of top-down knowledge which is the relations between Articles to further improve it. We test our hypothesis on several smaller subsets of the Wikipedia corpus and show that our proposed methodology leads to decent improvements in performance measures including Spearman's Rank correlation coefficient in most cases.


\keywords{Wikipedia Graph \and ESA \and Retrofitting.}
\end{abstract}
\section{Introduction}
Methods of capturing semantic relatedness in words/documents have been explored to a significant extent by the NLP community. Major approaches include data-driven learning of vector representations~\cite{glove,skipgram,globalcontext} from a very large corpora. While such \textit{bottom-up} approaches have proven to be effective in recent times, the outcomes of such models can be further amplified by using some form of \textit{top-down} knowledge gained over time, by humans. Such information has been crystallized into structured databases since as early as 1995~\cite{wordnet}. These top-down resources can be utilized to guide bottom-up expeditions to identify relations among words/documents.\\

Several works in this domain have surfaced recently. One such elegant idea, arrived at by Faruqui et al., is \textit{retrofitting}. We build our proposal on their paper~\cite{retro}, utilizing the technique rather unconventionally, in Explicit Semantic Analysis (ESA)~\cite{esa}, for improving word representations using top-down knowledge from Wikipedia.\\

Explicit Semantic Analysis is a popular technique  that represents the meaning of texts in a high-dimensional space of concepts derived from Wikipedia and using machine learning techniques explicitly represent the meaning of any text as a weighted vector of Wikipedia-based concepts. The information used here from the Wikipedia corpus are essentially the Article titles and the text or body of the Article. We suggest that by using an additional information which is often disregarded, such as the way in which one Wikipedia Article references to another, we could improve upon the earlier work as performed in ESA. One could imagine such references made back and forth between several Wikipedia Articles as a network of useful information. We incorporate this on top of ESA and see some minor improvements on several Benchmark tasks such as Word Similarity tasks like WS-353, RG-65 and MEN3k~\cite{ws,rg,men}. Especially we see improvements of around 2\% on average in performance measured using Spearman's Rank Correlation Coefficient. To represent the knowledge in a useful manner we take advantage of the Graph data structure, specifically an undirected Graph, wherein the nodes are the Articles and edges between nodes correspond to Inter-Relatedness between corresponding Articles. \\

In addition our work also comprises of implementing the paper - Retrofitting word vectors to semantic lexicons on a number of evaluation benchmarks across several data sets and cross-validating the results cited in the paper~\cite{retro}. In section 2 we discuss a few papers as background and related work that we consider and take motivation from, while forming our hypothesis. In section 3 we discuss our implementation of the paper that introduces Retrofitting~\cite{retro} which is a concept that we use to apply the Inter-Relatedness knowledge on top of ESA. In section 4 we introduce our main contributions and hypothesis to improve upon ESA using the discussed knowledge of Inter-Relatedness between Articles and present the results in doing so by drawing comparison before and after incorporating that knowledge.


\section{Background and Related Work}
\subsection{Retrofitting}
\textit{Faruqui et al.} introduced the novel method of retrofitting for capturing relational information among vector representations of words~\cite{retro}. Retrofitting is applied on word vectors as a \textit{post-processing} step such that, semantically related words will be nearer in their vector embedding space. This includes synonyms, hypernyms, hyponyms and paraphrases for the given word. Such relations among semantic lexicons are captured in resources including the Paraphrase Database~\cite{ppdb}, WordNet~\cite{wordnet} and FrameNet~\cite{framenet}.\\

Although retrofitting is agnostic to the manner in which the vectors were derived, a key point to note is the type of top-down knowledge database to be used for retrofitting. This in fact, depends on the nature of vectors at hand. That is, Faruqui et al. consider \textit{word vector} representations and hence any database which captures the many relations among \textit{words} would suffice (for example, WordNet, FrameNet). However if one were to refine the vector representations of documents for example, we need to consider a database of similar text documents to construct a graph wherein edges connecting two nodes correspond to two documents having similar content. Our proposal exploits this observation to build a customized graph of Wikipedia articles for retrofitting, as elaborated further in Section 4.

\subsection{Explicit Semantic Analysis (ESA)}
ESA is a novel method in which semantic relatedness between words or documents are captured by representation of meaning in a high-dimensional space of natural concepts derived from Wikipedia~\cite{esa}. Each concept is represented as an attribute vector of words that occur in the corresponding wikipedia article. Entries of these concept vectors are assigned weights using TF-IDF scheme. We should note that concepts are merely the titles of articles across which TF-IDF scores are calculated, for example, Computer Science, Automobile or India.\\

An inverted index is used to retrieve the relevant concepts for a given word. In the words of Markovitch et al., \textit{Each word appearing in the Wikipedia corpus can be seen as triggering each of the concepts it points to in the inverted index}~\cite{esa2}. Moreover, a piece of text (string of words) is simply represented as the centroid of the vectors representing its words in the concept space.

\subsection{Non-Orthogonal Explicit Semantic Analysis (NESA)}

One of the drawbacks in using ESA to obtain vector embeddings is that ESA inherently assumes that Wikipedia concepts are orthogonal to each other. Therefore, it considers that two words are related only if they co-occur in the same articles. However, two words can be related to each other even if they appear separately in related articles rather than co-occurring in the same articles. In this paper~\cite{nesa}, the authors come up with a technique to successfully overcome this issue leading to improvements in several of the Benchmarks and achieving state-of-the-art results.

To compute text relatedness, NESA uses relatedness between the dimensions of the distributional vectors to overcome the orthogonality in ESA model. This is accomplished by utilizing a square matrix $C_{n,n}$ where n is the total number of dimensions or concepts containing the correlation weights between the dimensions. While obtaining relatedness score between two words, instead of using something like a cosine similarity measure which is multiplying the two word vectors and normalizing them, they first multiply the first word vector with the square matrix $C_{n,n}$ and then later multiply this with the second word vector. What this essentially boils down to is that each concept dimension is made to further constitute a semantic vector built in another vector space. To construct $C_{n,n}$ consisting of correlation scores between Articles they provide four different methods and also discuss the performance improvements achieved by each one of them.

\section{Paper Implementation}
The objective of retrofitting is to reduce the Euclidean distance between a pair of \textit{similar} vectors and in doing so we end up with a richer representation of words using vectors. Therefore, we want our word vector in question $q_i$ to be close to the observed value $\hat{q_i}$ and close to its neighbours $q_j, \forall j$ such that $(i,j) \in E$, that is they constitute an edge in the graph constructed from lexicon database. Thus, the optimization criteria is~\cite{retro}: 
\begin{equation}
    \Psi(Q) = \sum_{i=1}^{n}\left(\alpha_i ||q_i - \hat{q_i}||^2 +\sum_{(i,j)\in E} \beta_{ij} ||q_i - q_j||^2 \right)
\end{equation}
$\alpha \ and \ \beta$ are hyper-parameters controlling the relative strengths of associations between two vectors\\
$n=$ Number of word vectors to be retrofitted  \\
Differentiating $\psi$ with respect to $q_i$ gives the following online update equation (refer paper by Faruqui et al. for details~\cite{retro}):
\begin{equation}
    q_i = \frac{\sum_{j:(i,j)\in E} \beta_{ij}q_j + \alpha_i\hat{q_i}}{\sum_{j:(i,j)\in E} \beta_{ij} + \alpha_i}
\end{equation}
Following the implementation details stated in~\cite{retro}, we apply retrofitting on 3 different sets of word vectors and by using 4 different semantic lexicon resources. We assess the performance of each combination with and without retrofitting on standard lexical semantic evaluation tasks. An analysis of the results obtained (see Table~\ref{tab4}) as well as comparison with the results reported in~\cite{retro} are detailed in Section 3.4. Refer Section 3.3 for details regarding evaluation tasks.
\subsection{Semantic Lexicons} We use Paraphrase Database (PPDB), WordNet$_{syn}$ (${WN}_{syn}$, captures only synonymous relations between words), WordNet$_{all}$ (${WN}_{all}$, captures relations including synonyms, hypernyms and hyponyms) and FrameNet \cite{ppdb,wordnet,framenet}. The size of the graphs obtained from the different top-down knowledge databases are shown in Table~\ref{tab1}.
\begin{table}
\centering
\caption{Size of graphs obtained from lexicon databases}\label{tab1}
\begin{tabular}{|l|c|c|}
\hline
Lexicon & Words & Edges\\
\hline
PPDB &  $102,90$ & $374,555$\\
WordNet$_{syn}$ & $148,730$ & $304,856$\\
WordNet$_{all}$ & $148,730$ & $934,705$\\
FrameNet & $10,822$ & $417,456$\\
\hline
\end{tabular}
\end{table}

\begin{table}
\centering
\caption{Word Vector Characteristics}\label{tab2}
\begin{tabular}{|l|c|c|}
\hline
Word Vector & \ Dimension \ & Training Corpus\\
\hline
Glove Vectors & $300$ & $6$ Billion tokens\\
Skip-Gram Vectors (SG) & $300$ & $100$ Billion tokens\\
Global Context Vectors (GC) \ & $50$ & $1$ Billion tokens\\
Multilingual Vectors (Multi) & $300$ & $100$ Billion tokens\\
\hline
\end{tabular}
\end{table}
\subsection{Word Vectors} We use publicly available pre-trained English word vectors for implementation. These include the Glove, Skip-Gram (SG) and Global Context (GC) vectors\cite{glove,skipgram,globalcontext}. The dimensions and training corpus characteristics for each set of word vectors are enlisted in Table~\ref{tab2}. We once again emphasise here that retrofitting is independent of the training models used for extracting vector representations~\cite{retro}.  

\subsection{Retrofitting: Evaluation Benchmarks}
The resultant vector representations are evaluated on tasks that test how well they capture both semantic and syntactic aspects of the representations along with an extrinsic sentiment analysis task. A summary of the different tasks are provided in Table~\ref{tab3}. For the sake of comparison of results, we perform the same evaluation tasks as given in the base paper~\cite{retro}.\\

We should note that the Syntactic Relations task is computationally expensive since for each list of 4 words (refer Section $3.4$), we find cosine similarity with each of the word types in the vector representations. Moreover, since the Skip-Gram vector set is enormous ($\approx$ 3GB compressed and $\approx$ 8GB in .txt file) the estimated time was large($\approx$ 22 hours) and we were facing issues with physical memory.
\begin{table}
\centering
\caption{Evaluation metrics and Dataset characteristics}\label{tab3}
\begin{tabular}{|l|c|c|}
\hline
\multicolumn{1}{|c|}{Evaluation Task} & Metric                                                                                              & Dataset                     \\ \hline
Word Similarity: MEN-3k               & \multirow{3}{*}{\begin{tabular}[c]{@{}c@{}}Spearman’s rank \\ correlation coefficient\end{tabular}} & 3000 word pairs             \\ \cline{1-1} \cline{3-3} 
Word Similarity: RG-65                &                                                                                                     & 65 word pairs               \\ \cline{1-1} \cline{3-3} 
Word Similarity: WS-353               &                                                                                                     & 353 word pairs              \\ \hline
Synonym Selection (TOEFL)             & \multirow{2}{*}{Accuracy}                                                                           & 80 sets of 5 words          \\ \cline{1-1} \cline{3-3} 
Sentiment Analysis (SA)               &                                                                                                     & 6920 + 872 + 1821 sentences \\ \hline
\end{tabular}
\end{table}
\subsubsection{Word Similarity}
We evaluate on 3 different benchmarks commonly used for measuring word similarity. These are the WS-353 dataset~\cite{ws} containing 353 pairs of English words that have been assigned similarity ratings by humans, the RG-65 dataset~\cite{rg} that contain 65 pairs of nouns and the MEN dataset~\cite{men} consisting of 3,000 word pairs. Assessing the similarity of words amounts to comparing the corresponding vectors using the cosine metric. We report Spearman’s rank correlation coefficient between the rankings produced by our model against the human rankings.
\subsubsection{Synonym Selection (TOEFL)}
The TOEFL synonym selection task is to select the semantically closest word to a target word from a list of four candidates~\cite{toefl}. The dataset contains 80 such questions. An example is “rug $\rightarrow$ {sofa, ottoman, carpet, hallway}”, with carpet being the most synonym-like candidate to the target word.

\subsubsection{Sentiment Analysis (SA)} 
Socher et al. created a treebank containing sentences annotated with labels on phrases and sentences from movie review excerpts~\cite{sentiment}. The coarse grained treebank of positive and negative classes has been split into training, development, and test datasets containing 6,920, 872, and 1,821 sentences, respectively. We train an L2-regularized logistic regression classifier on the average of the word vectors of a given sentence to predict the coarse-grained sentiment tag at the sentence level, and report the test-set accuracy of the classifier.

\subsubsection{Syntactic Relations (SYN-REL)}
The dataset is composed of analogous word pair that follow a common syntactic relation~\cite{synrel}. For example, walking and walked; The task is to find a word (word vector) $q_d$ that best fits the relation $q_a:q_b::q_c:q_d$, given $q_a,q_b,q_c$. An estimate of $q_d$ is obtained by using the vector offset method~\cite{skipgram,offset} as follows:
\begin{align}
    q_a:q_b &:: q_c:q_d\notag\\ 
    &\Rightarrow q_a - q_b = q_c - q_d\notag\\
    &\Rightarrow q_d = q_b – q_a + q_c\notag
\end{align}
Therefore, we find the vector q which has the highest cosine similarity to the estimate.  

\subsection{Results of Retrofitting}
In table~\ref{tab4}, the results of our implementation of retrofitting are presented. We present the absolute scores in each case of retrofitting. Note that the first three columns represent Spearman's rank correlation coefficient and the last two represent Accuracy. Note that all absolute values of performance measures are scaled to $100$ and reported as percentages. Key observations based on the model performance are enlisted below: 

\begin{table}[]
\centering
\caption{Results of absolute performance with retrofitting obtained by us; Spearman’s correlation (columns 2,3 and 4 from left) and accuracy (columns 5,6 and 7 from left) on different tasks. Higher scores are always better.}\label{tab4}
\begin{tabular}{|l|c|c|c|c|c|c|}
\hline
\textbf{Word Vector+Lexicon} & MEN-3k & RG-65 & WS-353 & TOEFL & SA    & SYN REL \\ \hline
\rowcolor[HTML]{C0C0C0} 
GLOVE                        & 73.75  & 76.95 & 60.85  & 88.75 & 78.31 & 78.1    \\ \hline
GLOVE+PPDB                  & 76.36  & 82.49 & 63.08  & 93.75 & 79.24 & 77.1    \\ \hline
GLOVE+${WN}_{syn}$                  & 73.71  & 79    & 61.34  & 93.75 & 78.97 & 66.8    \\ \hline
GLOVE+${WN}_{all}$                  & 75.94  & 85    & 61.52  & 91.25 & 79.24 & 66.1    \\ \hline
GLOVE+FN                     & 70.09  & 75.62 & 56.14  & 91.25 & 77.76 & 69.0    \\ \hline
\rowcolor[HTML]{C0C0C0} 
SG                           & 73.22  & 76.08 & 65.9   & 87.5  & 81.82 & 74.2       \\ \hline
SG+PPDB                      & 75.25  & 81.5  & 71.28  & 95    & 82.59 & 77.7      \\ \hline
SG+${WN}_{syn}$                      & 72.81  & 76.35 & 66.47  & 93.75 & 82.31 & 59.9      \\ \hline
SG+${WN}_{all}$                    & 73.65  & 84.39 & 67.4   & 93.75 & 81.82 & 59.1       \\ \hline
SG+FN                        & 69.88  & 79.02 & 61.84  & 87.5  & 81.38 & 64.64      \\ \hline
\rowcolor[HTML]{C0C0C0} 
GC                           & 31.4   & 62.99 & 62.58  & 57.5  & 67.38 & 18.8    \\ \hline
GC+PPDB                      & 40.31  & 68.99 & 64.58  & 71.25 & 68.92 & 27.3    \\ \hline
GC+${WN}_{syn}$                     & 34.87  & 69.3  & 63.18  & 63.75 & 68.53 & 13.8    \\ \hline
GC+${WN}_{all}$                    & 38.01  & 73.23 & 64.8   & 62.5  & 68.7  & 13.9    \\ \hline
GC+FN                        & 33.15  & 67.01 & 62.57  & 61.25 & 67.49 & 14.7    \\ \hline
\end{tabular}
\end{table}

\begin{itemize}
    \item Firstly we note that with similar hyperparameter settings, we obtain an equivalent performance corresponding to the results given in base paper~\cite{retro}.
    \item We obtain major improvements in word similarity tasks (columns 1-3 in table~\ref{tab4}) when implementing retrofitting with all the top-down knowledge databases with the exception of FrameNet. In fact, with the use of FrameNet on any evaluation task, we seldom observe any improvements (in some cases, we observe decrease in performance) with retrofitting. Faruqui et al. cite the reason being that the nature of top-down knowledge captured by FrameNet between words is very abstract and distantly related, (for example, \textit{prominent} and \textit{eye-catching}).     
    \item In the case of TOEFL task, we observe very high improvements in accuracy, the highest increment $(+13.5\%)$ occurring in the case of GC vectors with PPDB retrofit.
    \item We observe modest improvements in the case of Sentiment Analysis task (SA), the highest increment $(+1.5\%)$ occurring in the case of GC vectors with PPDB retrofit. 
    \item For the SYN-REL task, we observe improvement in case of GC $(+8.5\%)$ and SG $(+3.5\%)$ with PPDB retrofit. For all other cases, we always observe drop in accuracy, the highest reduction occurring in the case of SG vectors with ${WN}_{all}$ retrofit $(-15.1\%)$. As stated by Faruqui et al. in the base paper~\cite{retro}, the poor performance can be attributed to the fact that SYN-REL is inherently a syntactic task and by retrofitting, we only incorporate semantic information in the vector representations.
    \item Finally, we note that retrofitting with PPDB database gives the best improvement across a majority of the combinations (Word vector set $+$ Evaluation task), followed by ${WN}_{all}$, same as in the case of results provided in the base paper~\cite{retro}.
\end{itemize}
\section{Retrofitting in ESA}
We aim to incorporate the knowledge of Inter-Relatedness between Wikipedia Articles into ESA to improve the quality of vector representation of words/texts  and we use the method of retrofitting to accomplish this. Specifically, we retrofit the concept vectors which are vector embedding in the space of words using a graph like network consisting of the inter-relatedness knowledge. So, once we arrive at the representation of Wikipedia Articles as an attribute vector of words by using the TFIDF scheme, we incorporate semantic relatedness in vector embedding of Articles by means of an information which is the inter-relatedness of documents/articles in Wikipedia. \\

There might be cases where two words can be related to each other even if they appear separately  in related articles rather than co-occurring in the same articles which would then indicate that the word vectors have different representations and thus not similar. This is one of the drawbacks of ESA which we hope to overcome by incorporating our proposed methodology to enhance the word vectors obtained through ESA. Note that in the paper  Non-Orthogonal Explicit Semantic Analysis (NESA) \cite{nesa} the authors also try to overcome a similar problem but one of the key difference is that they accomplish this by making each concept dimension further constitute a semantic vector built in another vector space. Whereas we aim to solve this by retaining the concept space by using retrofitting as a means to enhance the vector representations using knowledge from Inter-Relatedness Graph obtained from Wikipedia corpus.\\

If we view an article in Wikipedia, in its description or content, there are links to other related Wikipedia articles in sections such as “References”, “See also” etc. and also as hyperlinks within the body of the Article. This type of information is particularly disregarded when working with plain text in ESA. We incorporate this form of knowledge so that closely interrelated concepts are nearer in their corresponding weighted representation of words. For the purpose of retrofitting, we build a graph-like data structure representing the inter related concepts explained in detail in Section 4.1. Then we modify the weighted representation of words for each concept to be nearer to itself as well as its related concepts using the same optimization criteria arrived at by Faruqui et al. (refer Equations 1 and 2).

\subsection{Wikipedia Inter-Relatedness Graph (Wiki Graph)}
The information that we are willing to retrofit the tf-idf matrix in ESA with is the inter relatedness between Wikipedia Articles. To do so, we construct an undirected graph with nodes as Article Titles and edges are constructed between two inter-related Articles. Inter-related articles could be viewed as the Top-Down Knowledge that we are incorporating into ESA while ESA is primarily a Bottom-Up approach. The main source of inter-relatedness is the hyperlinks present in each article which in Wikipedia parlance is called as internal links or wikilinks.

\begin{figure}[H]
    \centering
    \includegraphics[width=0.70\textwidth]{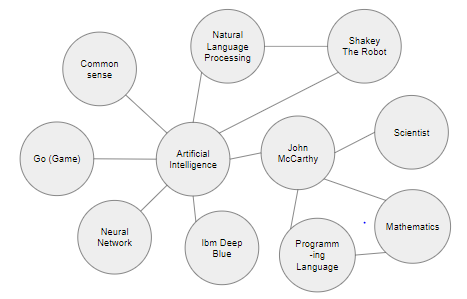}
    \caption{A snapshot of a tiny portion of Wikipedia inter-relatedness graph (undirected) showing nodes as Wikipedia Articles and edges between two related Articles.}
    \label{fig:1}
\end{figure}

There are two ways that we have come up with to generate the Graph. One is based on prior work by~\cite{wikilink} where they extract the hyperlinks from main text in each article which leads to another article within Wikipedia. The graph constructed this way is done so for the entire Wikipedia corpus at specific timestamps across history and we'll be using the latest one for our task in hand. Based on the corpus or the subset of the Wikipedia Articles that we are working with in ESA we extract the corresponding subgraph containing relations between only these Articles. Another method that we have come up with and is more efficient is directly parsing them from the raw data that we extract as dump from {\em Wikipedia\footnote{https://en.wikipedia.org/wiki/Special:Export}}. Each Article is represented as a json object and the key links\_out contains as value a list of all Wikipedia Articles that have been referred to from the original Article. Using this we reconstruct the graph from scratch specific to the dataset or corpus in hand. This is faster in computation time since we are constructing from base-up only what we need instead of having to reduce a much bigger graph. Note that for purposes of reusing the graph, we save them in the form of an adjacency list in text file in local storage.\\

Owing to space complexity, as discussed in later sections for storing the tf-idf matrix, we are working with a smaller corpus derived for each specific Evaluation task and the structure for the corresponding graphs is summarized in Table~\ref{graph_table}. A tiny portion of the graph is illustrated in Figure~\ref{fig:1} to show how Articles may be inter-related with each other.\\

\begin{table}[]
\centering
\caption{Summary of the extracted Graphs corresponding to each corpus that we use in ESA. Articles per word denotes the number of articles extracted to form the corpus for each word in the dataset corresponding to the Evaluation Task}\label{graph_table}
\begin{tabular}{|c|c|c|c|c|c|}
\hline
\textbf{Evaluation Task}         & \textbf{\begin{tabular}[c]{@{}c@{}}Articles \\ per word\end{tabular}} & \textbf{Articles} & \textbf{Nodes} & \textbf{Edges} & \textbf{\begin{tabular}[c]{@{}c@{}}\% Wikipedia \\ x$(10^{-2})$\end{tabular}} \\ \hline
\multirow{2}{*}{\textbf{WS-353}} & 5                                                                     & 2133              & 1459           & 4637           & 5.3325                                                                    \\ \cline{2-6} 
                                 & 6                                                                     & 2555              & 1768          & 5626           & 6.3875                                                                    \\ \hline
\multirow{2}{*}{\textbf{RG-65}}  & 10                                                                    & 498               & 150            & 179            & 1.245                                                                     \\ \cline{2-6} 
                                 & 20                                                                    & 993               & 338            & 445            & 2.4825                                                                    \\ \hline
\multirow{2}{*}{\textbf{MEN-3k}} & 3                                                                     & 2337              & 1775           & 8225           & 5.8425                                                                    \\ \cline{2-6} 
                                 & 5                                                                     & 3972              & 3026           & 13244          & 9.93                                                                      \\ \hline
\textbf{WikiLink Graphs}         & -                                                                     & 13,685,337        & 13,685,337     & 163,380,007    & -                                                                         \\ \hline
\end{tabular}
\end{table}

\subsection{Implementation}
In Appendix 1, we present a detailed example for implementation of our proposal. An overview is provided in the subsequent paragraphs. 

Firstly, the Wikipedia concepts are represented as vector embedding using words as dimensions with the TF-IDF scheme following the procedure for representing concepts in word space given in ~\cite{esa}. Then, the graph required for representing the inter-relatedness of Wikipedia concepts are extracted. (Section 5.1). 

The TF-IDF vectors for each concept are now retrofitted in the same fashion as in~\cite{retro} wherein the objective function (Eq. 1) is minimized thus resulting in the update step (Eq. 2).

Note that owing to the limited availability of compute resources at the time, the TF-IDF matrix for the entire Wikipedia corpus as well as an alternate (simple-Wikipedia) were unable to fit in the physical memory. Hence, we work on a much smaller corpora and extract a subset of Wikipedia pertaining to each Evaluation task (see Section 4.3) we consider at hand. To do so we take the top \textit{n} articles obtained by searching for each word used in the Evaluation dataset and then compiled them together.

\subsection{Evaluation}
We evaluate our model on word similarity tasks using the standard benchmarks WS-353, RG-65, and MEN-3k dataset~\cite{ws,rg,men}. A summary of vector characteristics along with performance of our model on the benchmarks are given in Table~\ref{results_proposal}.  
\subsubsection{Word Vectors} Owing to the computational constraints for testing our hypothesis on the entire Wikipedia corpus we derive much smaller subsets of Articles from Wikipedia separately for each corresponding Evaluation Benchmarks that we perform. For a given evaluation dataset, we extract the relevant concepts for each of the words present in the dataset from Wikipedia corpus using the Wikipedia library by uniformly extracting a fixed number of concepts for each word. We then construct our corpus wherein words are expressed in the space of the extracted concepts alone. For example, RG-65 (10) implies that for each word in RG-65 dataset top 10 concepts are extracted (See Column 1, Table~\ref{results_proposal}). Thus,in the resulting corpus, words are expressed in the $498$ dimensional space of concepts (Total extracted concepts = $2*65*10 = 1300$ of which $498$ are unique, see Column 2, Table~\ref{results_proposal}). As in ESA~\cite{esa}, we express a concept in terms of words occurring in corresponding Wikipedia article using TF-IDF scheme. Refer Columns 1-3, Table~\ref{results_proposal} regarding the vector characteristics used for evaluation.\\

Once we obtain the Word/Concept matrix, we apply retrofitting on concept vectors using the top-down knowledge graph specific to the evaluation dataset (refer Table~\ref{graph_table} and Section 4.1). Results of our model implementation are summarized in the next section.


\subsection{Results}
We ran experiments for each of the evaluation task coupled with it's characteristic corpus as shown in Table~\ref{graph_table} and report the Spearman's Rank correlation coefficient. To validate our hypothesis we compare the before and after results by first performing Explicit Semantic Analysis for each task followed by introducing inter-relatedness between Wikipedia Articles using Retrofitting. These are summarized in Table~\ref{results_proposal}. \\

Note that all absolute values of performance measures are scaled to $100$ and reported as percentages. Overall from the several Benchmarks that we ran we are able to show that on average using our method results in improvements in the performance measure. 
\begin{itemize}
\item For the WS-353 Benchmark, we observe an increase in performance by ~2\% and ~1\% for WS-353 (5) and WS-353 (6) variants where the number inside brackets denote the number of Articles used to form the corpus per word in the Benchmark dataset. 
\item We observe the highest improvement $(+3.01\%)$ in case of RG-65 (20) (RG-65 Benchmark with top 20 concepts extracted for each word).
\item Whereas, in the case of MEN-3k Evaluation task we observe a slight improvement $(+0.15\%)$ as well as a slight decrement $(-0.39\%)$ in performance compared to vanilla ESA for the MEN-3k (5) and MEN-3k (3) variants respectively. Hence, we can say that there is not enough evidence that our methodology provides performance improvements when measured using the MEN-3k Benchmark task. Furthermore for most of the tasks the performance increases by $1-2\%$ unless when there is no clear improvements.\\
\end{itemize}

Next, we compare our model performance against results obtained with Glove, Skip Gram (SG) and Global Context (GC) word vectors on word similarity tasks before and after retrofitting with different semantic lexicon databases (See Table~\ref{tab4} and refer Section 3.4). Note that we use a much smaller percentage of Wikipedia compared to the case of Glove and GC vectors owing to computational constraints (See Table~\ref{graph_table}) and has to be taken into account while comparing the results. 
\begin{itemize}
    \item On the MEN-3k benchmark, the highest correlation obtained by our methodology is $66.06\%$ (MEN-3k (3) $+$ Wiki Graph). Our results (ESA with Wiki Graph retrofit) are better than the performance of GC with/without retrofit by a margin of atleast $+22.55\%$. However, both SG and Glove vectors with/without retrofit give better results compared to our model (minimum and maximum margins of $+3.82\%$ and $+10.3\%$ respectively).
    \item Evaluating our model on RG-65 benchmark, the highest correlation obtained is $81.30\%$ (RG-65 (10) $+$ Wiki Graph). Once again, we produce better results with our model (ESA with Wiki Graph retrofit) than GC with/without retrofit by a margin of atleast $+8.07\%$. Performances of SG and Glove are similar to our model except in the case of retrofit with either PPDB or ${WN}_{all}$, wherein large improvements and higher correlations $(85\%)$ are observed (refer Section 3.4).
    \item On the WS-353 benchmark, we obtain $56.84\%$ correlation (WS-353 (5) $+$ Wiki Graph). With the exception of Glove $+$ FrameNet retrofit, all combinations of word vectors $+$ top-down database retrofit result in superior performance than our model (minimum and maximum decrements of $-4.01\%$ and $-14.44\%$ respectively). 
    
\end{itemize}

\begin{table}[]
\centering
\caption{Results of absolute performance obtained using ESA and with our proposed methodology. The metric for performance is Spearman’s Rank correlation coefficient (Higher scores are always better). Note that for a given evaluation task, we use different corpora. For example, in case of WS-353 (5), words are expressed in the $2133$ dimensional space of concepts (Column 2, Row 1 ), which is a consequence of constructing corpora by extracting only the top 5 concepts for each word.}\label{results_proposal}
\begin{tabular}{|c|c|c|c|c|}
\hline
\textbf{\begin{tabular}[c]{@{}c@{}}Evaluation \\ Task\end{tabular}} & \textbf{\begin{tabular}[c]{@{}c@{}}Dimension \\ (Word Vector )\end{tabular}} & \textbf{\begin{tabular}[c]{@{}c@{}}Dimension \\ (Concept Vector)\end{tabular}} & \textbf{\begin{tabular}[c]{@{}c@{}}Vector \\ Representation\end{tabular}} & \textbf{Performance} \\ \hline
\multirow{2}{*}{WS-353 (5)}                                         & \multirow{2}{*}{2133}                                                        & \multirow{2}{*}{162080}                                                        & ESA                                                                       & 54.61                \\ \cline{4-5} 
                                                                    &                                                                              &                                                                                & (+) Wiki Graph                                                             & 56.84                \\ \hline
\multirow{2}{*}{WS-353 (6)}                                         & \multirow{2}{*}{2555}                                                        & \multirow{2}{*}{183592}                                                        & ESA                                                                       & 55.12                \\ \cline{4-5} 
                                                                    &                                                                              &                                                                                & (+) Wiki Graph                                                             & 55.97                \\ \hline
\multirow{2}{*}{RG-65 (10)}                                         & \multirow{2}{*}{498}                                                         & \multirow{2}{*}{44401}                                                         & ESA                                                                       & 79.84                \\ \cline{4-5} 
                                                                    &                                                                              &                                                                                & (+) Wiki Graph                                                             & 81.30                \\ \hline
\multirow{2}{*}{RG-65 (20)}                                         & \multirow{2}{*}{993}                                                         & \multirow{2}{*}{67881}                                                         & ESA                                                                       & 78.09                \\ \cline{4-5} 
                                                                    &                                                                              &                                                                                & (+) Wiki Graph                                                             & 81.10                \\ \hline
\multirow{2}{*}{MEN-3k (3)}                                         & \multirow{2}{*}{2337}                                                        & \multirow{2}{*}{189661}                                                        & ESA                                                                       & 66.45                \\ \cline{4-5} 
                                                                    &                                                                              &                                                                                & (+) Wiki Graph                                                             & 66.06                \\ \hline
\multirow{2}{*}{MEN-3k (5)}                                         & \multirow{2}{*}{3972}                                                        & \multirow{2}{*}{253207}                                                        & ESA                                                                       & 62.86                \\ \cline{4-5} 
                                                                    &                                                                              &                                                                                & (+) Wiki Graph                                                             & 63.01                \\ \hline
\end{tabular}
\end{table}

\subsection{Conclusion \& Future Work}

Through our work in this project we first try to implement a paper conveying an elegant solution for enhancing Word Vectors by means of Retrofitting Semantic Lexicons to gain a deeper understanding on the subject and how research ideas are formed, hypothesized and evaluated. Then we propose our idea of improving the vector embeddings obtained through Explicit Semantic Analysis by introducing top-down knowledge of Inter-Relatedness between Wikipedia Articles. We do so by constructing this knowledge using the network of hyperlinks in each Article linking other similar Articles to itself and formally storing it as an undirected graph with nodes as Articles and Edges showing Inter-relatedness between two such nodes.\\

Based on our experimentation with different datasets, each evaluated using different Benchmarks, we show that this methodology in fact leads to decent improvements in performance of the Word vectors obtained after introducing this top-down knowledge by means of Retrofitting to ESA. In the future this could be extended to much larger dataset like the entire Wikipedia corpus to possibly realize state of the art results since we had to limit ourselves to smaller subsets of Wikipedia corpus owing to the computational constraints present at the moment.


\section{Appendix 1}
In this section, we provide a step by step walkthrough for implementing our proposal (refer Section 4). Consider the piece of text, "Information Retrieval in Search Engines". After pre-processing the query we are left with the following list of tokens:
\begin{center}
    ['information',  'retrieval',  'search',  'engines']
\end{center}
In short, we are left with 4 word tokens:
\begin{center}
    [$w_1$, $w_2$, $w_3$, $w_4$]
\end{center} 
In ESA~\cite{esa}, we represent each of these words as vectors in the concept space. We first retrieve the concepts corresponding to the words using inverted indexing. We then represent the concepts themselves as a vector embedding of the words which occur in the article corresponding to the concept (concept is just the title of article).
\begin{align}
    w_1 &\rightarrow doc_1, \ doc_2, \ doc_3 \notag\\
    w_2 &\rightarrow doc_1, \ doc_2 \notag\\
    w_3 &\rightarrow doc_1, \ doc_3 \notag\\
    w_4 &\rightarrow doc_2, \ doc_3 \notag
\end{align}
Thus, $w_2$ occurs in ${doc}_1$, ${doc}_2$, words $w_1$,$w_2$,$w_3$ occur in ${doc}_1$ and so on. This relationship is captured in the word/concept matrix (refer~\ref{wcmatrix}).


\begin{table}[]
\centering
\caption{\textbf{Word/Concept Matrix}. Concepts can be expressed in 4 dimensional space of words occuring in corresponding articles. Consequently, words can be expressed in 3 dimensional space of concepts.}\label{wcmatrix}
\begin{tabular}{|
>{\columncolor[HTML]{FFFFFF}}c |
>{\columncolor[HTML]{EFEFEF}}c |
>{\columncolor[HTML]{EFEFEF}}c |
>{\columncolor[HTML]{EFEFEF}}c |}
\hline
{\color[HTML]{000000} \textbf{Word/Concept}} & \cellcolor[HTML]{FFFFFF}{\color[HTML]{000000} \textbf{$c_1$}} & \cellcolor[HTML]{FFFFFF}{\color[HTML]{000000} \textbf{$c_2$}} & \cellcolor[HTML]{FFFFFF}{\color[HTML]{000000} \textbf{$c_3$}} \\ \hline
{\color[HTML]{000000} \textbf{$w_1$}}           & {\color[HTML]{000000} ${tf}_{1}^{1}$}                                  & {\color[HTML]{000000} ${tf}_{1}^{2}$}                                  & {\color[HTML]{000000} ${tf}_{1}^{3}$}                                  \\ \hline
{\color[HTML]{000000} \textbf{$w_2$}}           & {\color[HTML]{000000} ${tf}_{2}^{1}$}                                  & {\color[HTML]{000000} ${tf}_{2}^{2}$}                                  & {\color[HTML]{000000} $0$}                                  \\ \hline
{\color[HTML]{000000} \textbf{$w_3$}}           & {\color[HTML]{000000} ${tf}_{3}^{1}$}                                  & {\color[HTML]{000000} $0$}                                  & {\color[HTML]{000000} ${tf}_{3}^{3}$}                                  \\ \hline
{\color[HTML]{000000} \textbf{$w_4$}}           & {\color[HTML]{000000} $0$}                                  & {\color[HTML]{000000} ${tf}_{4}^{2}$}                                  & {\color[HTML]{000000} ${tf}_{4}^{3}$}                                  \\ \hline
\end{tabular}
\end{table}

The concept vectors $\vec{c_1}, \ \vec{c_2}, \ \vec{c_3}$ corresponding to ${doc}_1, \ {doc}_2, \ {doc}_3$ respectively are vectors of words:
\begin{align}
    \Vec{c_1} &= {tf}_{1}^{1} \ \hat{w_1} + {tf}_{2}^{1} \ \hat{w_2} + {tf}_{3}^{1} \ \hat{w_3} \notag\\
    \Vec{c_2} &= {tf}_{1}^{2} \ \hat{w_1} + {tf}_{2}^{2} \ \hat{w_2} + {tf}_{4}^{2} \ \hat{w_4} \notag\\
    \Vec{c_3} &= {tf}_{1}^{3} \ \hat{w_1} + {tf}_{3}^{3} \ \hat{w_3} + {tf}_{4}^{3} \ \hat{w_4} \notag
\end{align}
$c_i \implies i^{th} \ concept$\\
$w_i \implies i^{th} \ word $\\
$tf_{i}^{j} \implies TF-IDF \ of \ i^{th} \ word \ occurring \ in \ j^{th} \ document$\\

\noindent Applying retrofitting (using Eqn. 1 and 2 from Section 3) to these concept vectors based on the Wikipedia graph, we get,\\
\begin{align}
    \Vec{c_1} &= {{tf}_{1}^{1}}^R \ \hat{w_1} + {{tf}_{2}^{1}}^R \ \hat{w_2} + {{tf}_{3}^{1}}^R \ \hat{w_3} \notag\\
    \Vec{c_2} &= {{tf}_{1}^{2}}^R \ \hat{w_1} + {{tf}_{2}^{2}}^R \ \hat{w_2} + {{tf}_{4}^{2}}^R \ \hat{w_4} \notag\\
    \Vec{c_3} &= {{tf}_{1}^{3}}^R \ \hat{w_1} + {{tf}_{3}^{3}}^R \ \hat{w_3} + {{tf}_{4}^{3}}^R \ \hat{w_4} \notag
\end{align}
${tf_{i}^{j}}^R \implies TF-IDF \ of \ i^{th} \ word \ occurring \ in \ j^{th} \ document \ after \ retrofit$\\
For example, ${{tf}_{1}^{3}}^R$ is the vector weight of $w_3$ occurring in $doc_1$ after retrofit.\\

Ultimately, we arrive at the word vectors $\Vec{w_1}, \Vec{w_2}, \Vec{w_3}, \Vec{w_4}$ which are vector embedding in the concept space, where the weight (magnitude) corresponding to concept $\Vec{c_j}$ for $i^{th}$ word vector $\Vec{w_i}$ is given by ${tf_{i}^{j}}^R$. Thus,
\begin{align}
    \Vec{w_1} &= {{tf}_{1}^{1}}^R \ \hat{c_1} + {{tf}_{1}^{2}}^R \ \hat{c_2} + {{tf}_{1}^{3}}^R \ \hat{c_3} \notag\\
    \Vec{w_2} &= {{tf}_{2}^{1}}^R \ \hat{c_1} + {{tf}_{2}^{2}}^R \ \hat{c_2} \notag\\
    \Vec{w_3} &= {{tf}_{3}^{1}}^R \ \hat{c_1} + {{tf}_{3}^{3}}^R \ \hat{c_3} \notag\\
    \Vec{w_4} &= {{tf}_{4}^{2}}^R \ \hat{c_2} + {{tf}_{4}^{3}}^R \ \hat{c_3} \notag
\end{align}
Therefore, the resultant vector embedding in the concept space for "Information Retrieval in Search Engines" is given by,
\begin{align}
    \Vec{W} &= \Vec{w_1} + \Vec{w_2} + \Vec{w_3} + \Vec{w_4}\notag\\
    \Vec{W} &= ({{tf}_{1}^{1}}^R + {{tf}_{2}^{1}}^R + {{tf}_{3}^{1}}^R) \hat{c_1} \notag\\ &+ ({{tf}_{1}^{2}}^R + {{tf}_{2}^{2}}^R + {{tf}_{4}^{2}}^R) \hat{c_2} \notag\\ 
    &+ ({{tf}_{1}^{3}}^R + {{tf}_{3}^{3}}^R + {{tf}_{4}^{3}}^R) \hat{c_3} \notag
\end{align}

%
%
%
%

\end{document}